\documentclass[10pt,twocolumn,letterpaper]{article}

\usepackage{wacv}              % To produce the CAMERA-READY version
\usepackage{graphicx}
\usepackage{amsmath}
\usepackage{amssymb}
\usepackage{booktabs}
\usepackage{algorithm}
\usepackage{algpseudocode}% http://ctan.org/pkg/algorithmicx

\usepackage[pagebackref,breaklinks,colorlinks]{hyperref}

\newcommand{\RomNum}[1]{\MakeUppercase{\romannumeral #1}}

\usepackage[capitalize]{cleveref}
\crefname{section}{Sec.}{Secs.}
\Crefname{section}{Section}{Sections}
\Crefname{table}{Table}{Tables}
\crefname{table}{Tab.}{Tabs.}

\def\wacvPaperID{1247} % *** Enter the WACV Paper ID here
\def\confName{WACV}
\def\confYear{2025}

\begin{document}

\title{Now You See Me: Context-Aware Automatic Audio Description}

\author{Seon-Ho Lee\thanks{Work done during an internship at Amazon Prime Video.}\\
Korea University\\
% {\tt\small seonholee@mcl.korea.ac.kr}
\and
Jue Wang\\
Amazon AGI\\
% {\tt\small juewangn@amazon.com}
\and
David Fan\thanks{Work done while at Amazon Prime Video.}\\
Meta FAIR\\
% {\tt\small davidfan@meta.com}
\and
Zhikang Zhang\\
Amazon AGI\\
% {\tt\small zhikang@amazon.com}
\and
Linda Liu\\
Amazon Prime Video\\
% {\tt\small lindliu@amazon.com}
\and
Xiang Hao\\
Amazon Prime Video\\
% {\tt\small lindliu@amazon.com}
\and
Vimal Bhat\\
Amazon Prime Video\\
\and
Xinyu Li\\
Amazon AGI\\
}
\maketitle

%%%%%%%%% ABSTRACT
\begin{abstract}

Audio Description (AD) plays a pivotal role as an application system aimed at guaranteeing accessibility in multimedia content, which provides additional narrations at suitable intervals to describe visual elements, catering specifically to the needs of visually impaired audiences. In this paper, we introduce $\mathrm{CA^3D}$, the pioneering unified Context-Aware Automatic Audio Description system that provides AD event scripts with precise locations in the long cinematic content. Specifically, $\mathrm{CA^3D}$ system consists of: 1) a Temporal Feature Enhancement Module to efficiently capture longer term dependencies, 2) an anchor-based AD event detector with feature suppression module that localizes the AD events and extracts discriminative feature for AD generation,  and 3) a self-refinement module that leverages the generated output to tweak AD event boundaries from coarse to fine. Unlike conventional methods which rely on metadata and ground truth AD timestamp for AD detection and generation tasks, the proposed $\mathrm{CA^3D}$ is the first end-to-end trainable system that only uses visual cue. Extensive experiments demonstrate that the proposed  $\mathrm{CA^3D}$ improves existing architectures for both AD event detection and script generation metrics, establishing the new state-of-the-art performances in the AD automation.
\end{abstract}

%%%%%%%%% BODY TEXT

\section{Introduction}
\label{sec:intro}

Movie audio description (AD) is the verbal narration which describes the visual elements in the movie. Movie AD aims to enable people to understand the story of the movie only with the sounds. Hence, it is especially important for that visually impaired people can have the equal opportunity to enjoy movies. However, the manual generation of AD is time-consuming and expensive; up to $\$75$ per minute of content~\footnote{https://www.3playmedia.com/blog/how-much-does-audio-description-cost/}. To reduce the enormous cost for AD generation, various techniques~\cite{han2023autoad, han2023autoad2} have been proposed. However, the automated AD generation is still challenging and requires more improvements for practical usage.

% plays a crucial role in streaming media by providing audiences with visual content in an auditory format, delivering an experience akin to watching the content itself. This not only ensures inclusively for individuals with disabilities, allowing them to engage in our visual culture, but also opens doors to enriched experiences like auditory movies or condensed movie summaries~\cite{perego2016gains}. However, the manual creation of AD comes at a high cost, with estimates suggesting up to $\$75$ per minute of content~\footnote{https://www.3playmedia.com/blog/how-much-does-audio-description-cost/}. Hence, many techniques for the automated AD generation have been proposed to reduce the cost. 
% However, automated AD generation remains a challenging frontier in research, yet to be fully explored and perfected. Automated AD system is required.

\begin{figure}[t]
    \centering
    \includegraphics[width=0.99\linewidth]{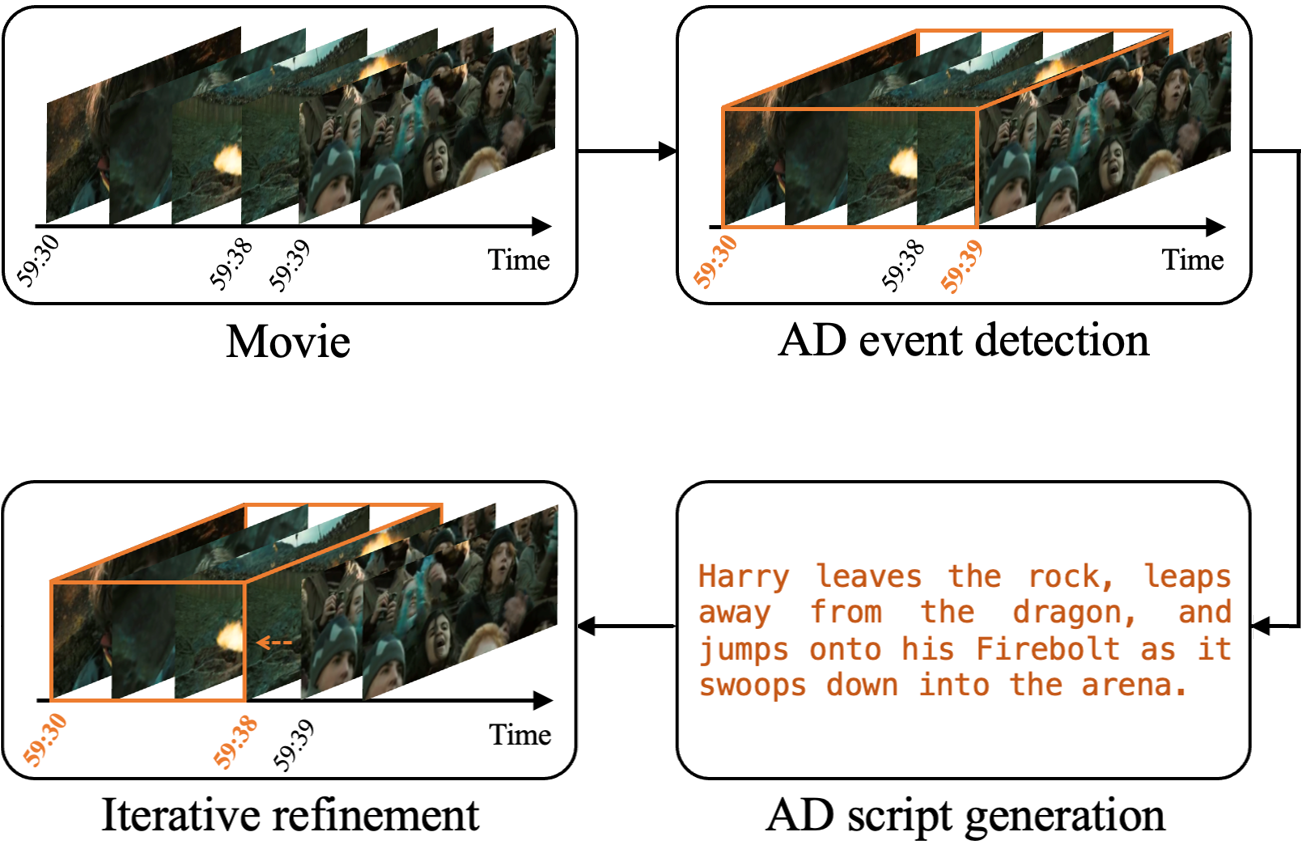}
    \caption{$\mathbf{CA^3D}$: We propose a one-stage AD Automation system that detects AD events, generates AD scripts and refines AD detection on a whole movie. } 
    \label{Fig1:example}
\end{figure}

To build an automated AD system for movies, two sub-problems should be addressed: AD event detection and AD script generation. Given a movie, AD event detection aims to find the temporal locations where AD should be provided while the AD script generation aims to create textual description for the detected AD event window. Compared to the standard (dense) video captioning~\cite{iashin2020multi, krishna2017dense, wang2018bidirectional, wang2020event}, the AD automation is more challenging because the boundary between AD and non-AD events are ambiguous. Hence, it requires long-term context from movies to decide when and where to generate AD. Despite of its importance, there is no previous work addressing the AD automation properly. 

Recently, AutoAD~\cite{han2023autoad} and its improved version, AutoAD~\RomNum{2}~\cite{han2023autoad2} have been proposed to tackle the AD automation. However, they decouple the AD detection and generation as two separate tasks and simplify each of them. Specifically, AutoAD~\RomNum{2} leverages subtitle information as the context to exclude non-AD events and predicts if at least one AD event exists in the speech gap, which cannot produce precise AD timestamp and may be biased on the data statistic. For the AD generation, conventional methods~\cite{han2023autoad, han2023autoad2} assume the access to the ground truth AD events locations. To follow the story more accurately, they also leverage the additional information such as the previous AD scripts, subtitles, and character data to generate AD scripts. However, such metadata is not always available for all movies. Moreover, the AD detection and generation should be coupled as a whole system in the real industry, where has no access to any prior. 

In this paper, we propose an end-to-end trainable system to achieve the context-aware automatic audio description ($\mathrm{CA^3D}$), which is the first unified algorithm to both AD event detection and AD script generation. Specifically, we first propose a temporal feature enhancement module to capture longer term dependencies to expand the temporal horizon of input by employing the structured state space sequence (S4) model, and then we introduce an anchor-based AD event detector associated with a feature suppression module which inhibits the following AD generator from using the information irrelevant to the AD event. At last, we also propose a self-refinement module as an option for fine-tuning the AD event location and scripts. Through extensive experiments, we demonstrate the superior performance of our proposed system on academia public benchmark (MADv2~\cite{han2023autoad2,soldan2022mad} dataset). We summarize the contribution of this paper as following:

\noindent $\bullet$ We propose the first unified AD automation system, $\mathrm{CA^3D}$, which generates AD scripts with precise timestamp on the entire movie. 

\noindent $\bullet$ We first employ the S4 model as a temporal feature enhancement module and sub-sequentially propose anchor-based detector, differentiable feature suppression module and an optional self-refinement module to work seamlessly in the AD automation system.

% \item We develop an anchor-based AD detector with a differentiable feature suppression module that facilitates the extraction of discriminative visual features for the AD generation.

% \item We propose a self-refinement module as an add-on in the $\mathrm{CA^3D}$, which further improves the AD detection and generation from coarse to fine. 

\noindent $\bullet$ We achieve the promising performances on both AD detection and generation tasks. Notably, even without leveraging external data and ground truth locations, the proposed algorithm shows competitive or better results to the previous methods which exploit those additional information.

\section{Related Work}
\label{sec:related_work}
\subsection{Dense Video Captioning} 
The dense video captioning (DVC) is one of the closest applications to the AD automation, both necessitating the detection of events within an untrimmed video and the subsequent generation of descriptive content for each identified event. Previous research efforts can be broadly categorized into two groups: 1) two-stage models such as those presented in works like \cite{iashin2020multi, krishna2017dense, wang2018bidirectional, wang2020event}, which bifurcate the task into event detection and trimmed video captioning; 2) joint models exemplified by \cite{deng2021sketch, chen2021towards, li2018jointly, mun2019streamlined}, which concurrently optimize detection and generation tasks by exploiting cross-modal alignment and events connection. While the workflow of DVC and AD automation shares similarities, the latter is more challenging due to the fact that AD automation necessitates capturing significantly longer contextual information from movies, in contrast to the average video length of 150 seconds in the Activity Net Captions \cite{krishna2017dense}, which serves as a de facto benchmark for DVC. Unlike video captions, AD automation engages in auditory story understanding, imposing supplementary requirements for both location and content considerations.

\begin{figure*}[t]
    \centering
    \includegraphics[width=\linewidth]{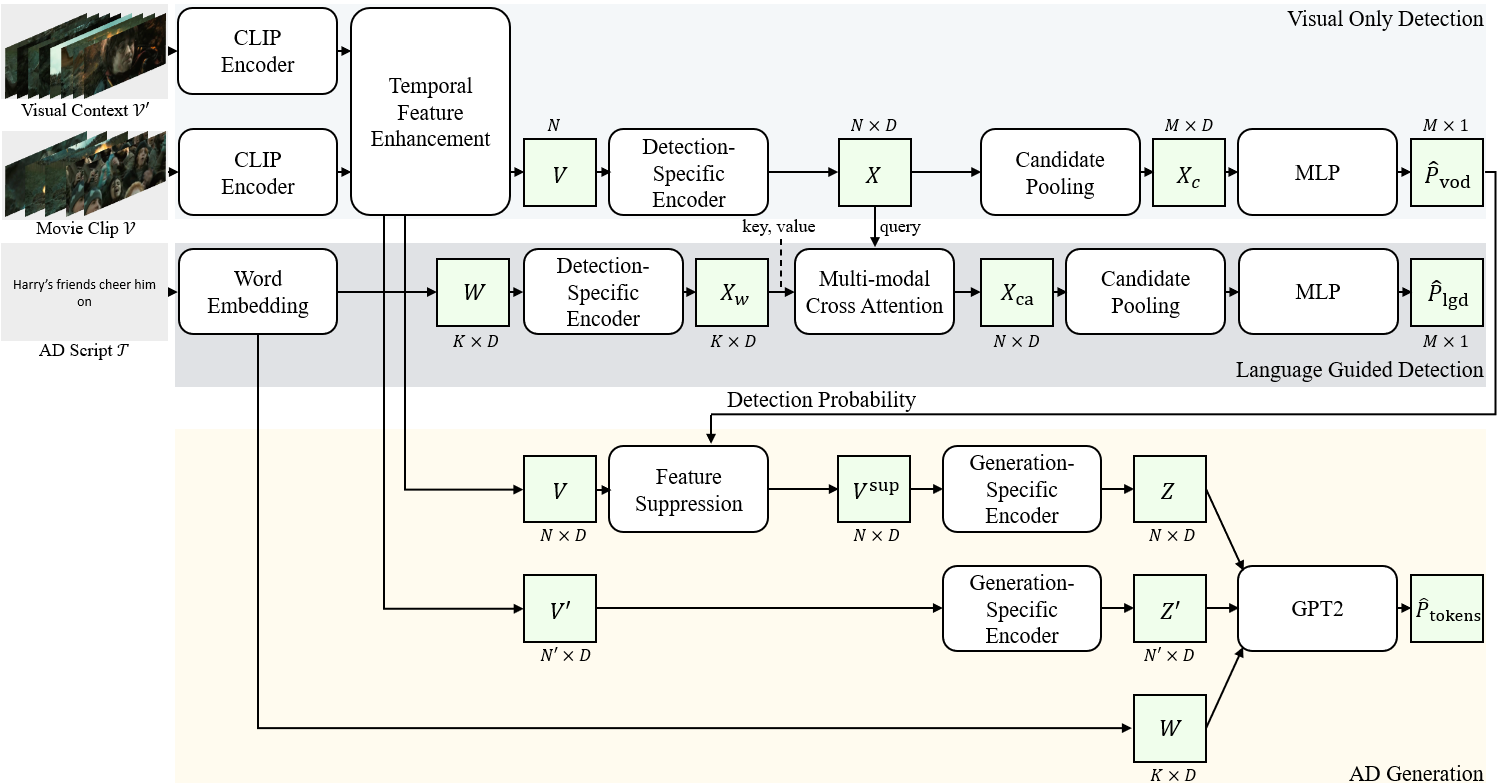}
    \caption{An overview of $\mathrm{CA^3D}$ network. The upper part shows the detector architecture and the bottom part depicts the generator architecture.}
    \vspace*{-0.1cm}
    \label{fig:overview}
    \vspace*{-0.1cm}
\end{figure*}

\subsection{Long-form Video Understanding with S4} 
As mentioned above, the AD automation needs to capture the long-term dependencies from movies. In the long-form video understanding, there are two major challenges concluded from the previous researches~\cite{sunlong2022,bertasius2021space, islam2022long, lvu2021,wu2022memvit, wang2022deformable}: efficiency and effectiveness. Efficiency issue comes from the large memory and computational cost of long input while the effectiveness challenge represents how to learn discriminative feature from redundant video sequences. To tackle these challenges, 
% LF-VILA~\cite{sunlong2022} develops a hierarchical feeding architecture to include more frames in the model, thus capturing longer temporal information. Similarly, MeMViT~\cite{wu2022memvit} better utilizes temporal information by emerging the previously cached ``memory" from the past. The pyramid structure leveraged by LF-VILA and MeMViT shows efficiency improvements, but may lose low-level spatial-temporal contextual information. 
Gu et al.~\cite{gu2021efficiently} proposed a structured state-space sequence model, a novel alternative to CNNs and transformers, which models the long-range dependencies by simulating a linear time invariant (LTI) system. Subsequently, S4ND~\cite{nguyen2022s4nd} and ViS4mer~\cite{islam2022long} extend S4 model to the video classification task. Finally, Wang et al.~\cite{wang2023selective} further improve the efficiency of S4 model with additional selective module formulating the S5 model. In contrast to previous applications of the S4 model that leverage it to model long sequential inputs, we uniquely employ the S4 module as a visual enhancement encoder. Our approach aims to distill longer-term memory into visual features from shorter clips. Notably, our design differs from LSTCL~\cite{wang2022long}, LSMCL~\cite{wang2022deformable}, and BraVe~\cite{recasens2021broaden} by abandoning the dual-encoder with symmetric contrastive learning. This departure significantly improves efficiency and practicality.

\subsection{Audio Description Automation} 
There are initial explorations~\cite{shen2023fine,soldan2022mad, han2023autoad, han2023autoad2} that try to generate AD and predict AD locations, but the current solutions are still far from being practically useful to scale up the AD automation. Starting from the data curation, LSMDC~\cite{rohrbach2017movie}, M-VAD~\cite{torabi2015using}, QuerYD~\cite{oncescu2021queryd}, and MPII-MD~\cite{rohrbach2015dataset} gather linguistic information from movies to curate clip-level video captioning task. The size of these datasets are either small-scaled~\cite{rohrbach2017movie,torabi2015using, rohrbach2015dataset} or not from the cinematic data~\cite{oncescu2021queryd}. To improve these, Soldan et al.~\cite{soldan2022mad} propose the MAD dataset which is a large-scaled benchmark with cinematic content on visual grounding task. Based on which, Han et al.~\cite{han2023autoad} propose a cleaner version, named MADv2 dataset, and introduce AutoAD for AD generation on trimmed AD events. Followed by AutoAD~\cite{han2023autoad}, AutoAD~\RomNum{2}~\cite{han2023autoad2} includes a new temporal segment proposing module indicating whether or not a AD should be generated within a speech gap. In addition, this work also improves the generation part with a Flamingo-style~\cite{alayrac2022flamingo} architecture to generate better AD scripts. However, prior works decouple the AD detection and generation as two separate tasks and each task is simplified. For example, AutoAD~\RomNum{2}~\cite{han2023autoad2} only provides a binary decision within a speech gap and the duration of the gap is prefixed based on the statistic of MADv2 dataset~\cite{soldan2022mad}. Moreover, both AutoAD~\cite{han2023autoad} and AutoAD~\RomNum{2}~\cite{han2023autoad2} generate AD scripts on the ground truth AD locations, which is impractical in the real world scenarios. Thus, we propose an unified AD automation system in this paper, that can automatically detect and generate AD on cinematic data at scale.

\section{Method}
\label{sec:alg}
% We start by the problem definition of the AD automation system($\S$~\ref{sec:alg-1}), followed by the novel Context-Aware Feature Enhancement module with S4 model($\S$~\ref{sec:alg-2}). Then, we introduce our proposed AD detection module ($\S$~\ref{sec:alg-3}) and connected AD script generation module ($\S$~\ref{sec:alg-4}). Lastly, we provide the details of our proposed loss function ($\S$~\ref{sec:alg-5})
% and integrative refinement mechanism ($\S$~\ref{sec:alg-6}).

\subsection{Problem Definition}
\label{sec:alg-1}
Given a movie clip ${\cal V} = \{I_1, I_2, \ldots, I_N\}$ with $N$ consecutive frames and its visual context ${\cal V}'$, the proposed $\mathrm{CA^3D}$ first enriches ${\cal V}$ with ${\cal V}'$. From $\cal V$ and ${\cal V}'$, we obtain the enhanced visual feature $V$ by using the image encoder and the feature enhancement module. Then, $V$ is sent to the AD detector\footnote{We assume only one AD event within $\cal V$ during the training.} to predict $\{y_1, y_2, \ldots, y_N\}$, where $y_i \in \{0, 1\}$ indicates whether the $i$-th frame $I_i$ belongs to AD events. Followed by the detection results, a feature suppression module is applied to extract AD related representation: $V_\mathrm{sup}$ from $V$. Lastly, the AD script ${\cal T} = \{t_1, t_2, \ldots t_K\}$ of $K$ words is generated, which describes the content in the way that people can enjoy the movie by hearing the Text To Speech (TTS)\footnote{Please note the TTS generation is out of scope of this work. This work only focuses on AD detection and generation.} of it.

% detects the temporal location of the AD event within $\cal V$\footnote{We assume only one AD event within $\cal V$ during the training.}. In other words, it predicts $\{y_1, y_2, \ldots, y_N\}$, where $y_i \in \{0, 1\}$ indicates whether the $i$-th frame $I_i$ belongs to AD events. Followed by the detection results, the AD script ${\cal T} = \{t_1, t_2, \ldots t_K\}$ of $K$ words (or tokens) are generated, describing the content in the way that people can still enjoy the movie by hearing the Text To Speech (TTS)\footnote{Please note the TTS generation is out of scope of this work. This work only focuses on AD detection and generation.} of AD scripts.

% \begin{figure*}[t]
%     \centering
%     \includegraphics[width=0.99\linewidth]{figure/generator.png}
%     \caption{An overview of generator architecture.}
%     %\vspace*{-0.1cm}
%     \label{fig:generator}
%     %\vspace*{-0.1cm}
% \end{figure*} 

\subsection{Temporal Feature Enhancement Module}
As observed in previous researches~\cite{han2023autoad,han2023autoad2}, both the location and content of AD events adhere to the story line, necessitating the model to accurately grasp long-term reasoning. Therefore, conventional algorithms~\cite{han2023autoad,han2023autoad2} leverage meta data, such as subtitle or character bank, to exploit context information in the AD automation. However, such information may not be available for all movies which makes it difficult to scale up. On the other hand, recent study on the S4 model~\cite{gu2021s4, islam2022long} has shown its superior performance in modeling long-form video, with the linear complexity to the input length. Compared to the transformers and RNNs, S4 model can capture longer history with cheaper cost.

\label{sec:alg-2}
\noindent\textbf{Preliminaries -- S4 Model:}
We start from the state-space model,~\textit{i.e.}, a linear time invariant system, which can be written as:
\begin{align}
\begin{split}
\label{SSM}
{x}^\prime (t) &= {A}{x}(t)+{B}{u}(t) \\
{y}(t) &= {D}{x}(t) + {E}{u}(t).
\end{split}
\end{align}

This formulation simply projects an input signal ${u}(t)$ from one-dimensional space to an $\textrm{N}$-dimensional latent space ${x}(t)$, which is then mapped back to a one-dimensional output signal ${y}(t)$. To implement~\eqref{SSM} with discrete inputs like image/word tokens, it can be discredited by using a learnable step size $\Delta$~\cite{gu2021combining}, which can be rewritten as:
\begin{align}
\begin{split}
\label{S4}
&{x}_k = \Bar{{A}}{x}_{k-1}+\Bar{{B}}{u}_k \\
&{y}_k=\Bar{{D}}{x}_k,
\end{split}
\end{align}
where $\Bar{{A}}=({I}+\frac{\Delta\cdot {A}}{2})/({I}-\frac{\Delta\cdot {A}}{2})$, $\Bar{{B}}=\Delta\cdot {B}/(I-\frac{\Delta\cdot {A}}{2})$, $\Bar{{D}} = {D}$ and ${E}$ can be replaced by residual connection. Furthermore,~\eqref{S4} can be solved using a discrete convolution~\cite{gu2021efficiently}:
\begin{equation}
\label{S4_conv}
    {y} = \Bar{{\cal K}} \circledast {\cal U},
\end{equation}
where ${\cal U}=\{u_0, u_1,\dots, u_{k-1}, u_k\}$ and $\Bar{{\cal K}} = \{\Bar{{D}}\Bar{{B}}, \Bar{{D}}\Bar{{A}}\Bar{{B}},\dots,\Bar{{D}}\Bar{{A}}^{\textrm{L}-1}\Bar{{B}}\}$ is a discredited convolutional kernel and $\textrm{L}$ is the sequence length. 

It is found in \cite{gu2021efficiently} that $\Bar{\cal K}$ can become a closed-form expression if the matrix $A$ becomes diagonal and low-rank (structured by the HiPPO theory~\cite{gu2020hippo}). As a result,~\eqref{S4_conv} is linear to the input length and can be efficiently computed using fast Fourier transform (FFT) and inverse FFT, without multiplying the matrix $A$ by $L-1$ times\footnote{Please refer to~\cite{gu2020hippo} for more details and relevant proofs.}. This advantage shapes the S4 model as an efficient architecture capturing long temporal dependencies. 

\noindent\textbf{Context-aware Temporal Feature Enhancement:} Since AD involves storytelling, the context of the movie is crucial for both AD detection and generation; For instance, to generate the AD script for `Dooku fires \textit{again}' in Star Wars, it is essential to understand Dooku's actions in the preceding scenes. Therefore, we incorporate visual context, represented by the movie clip ${\cal V}'=\{I'_1,\ldots,I'_{N'}\}$ with $N'$ frames, immediately preceding the movie clip $\cal V$. Subsequently, we enhance the features from both ${\cal V}'$ and $\cal V$ by conveying context information through the feature enhancement module $h_\mathrm{S4}$, which comprises S4 layers:
\begin{equation} \label{eq:enhancement}
    [V', V] = h_\mathrm{S4}(h_\mathrm{CLIP}({\cal V}'), h_\mathrm{CLIP}({\cal V})),
\end{equation}
Here, $h_\mathrm{CLIP}$ represents the CLIP image encoder~\cite{radford2021clip}\footnote{We opt for the CLIP image encoder because the MADv2~\cite{soldan2022mad} dataset provides only frame-level CLIP-encoded features.}. $V' = [v'_1, v'_2, \ldots, v'_{N'}] \in \mathbb{R}^{N' \times D}$ and $V = [v_1, v_2, \ldots, v_N] \in \mathbb{R}^{N \times D}$ are the enhanced feature maps of the visual context ${\cal V}'$ and movie clip $\cal V$. Leveraging the robust temporal capacity of the S4 model~\cite{gu2021s4, gu2021efficiently, islam2022long}, the proposed feature enhancement module $h_\mathrm{S4}$ distills long-term dependencies into the subsequent visual features. Consequently, the enhanced movie clip feature $V$ (length of $N$) encapsulates visual context information from $N+N'$ frames. In Section~\ref{sec:exp}, we demonstrate the superior performance of the temporally enhanced feature map.

% First, we map each frame in $\cal V$ to a feature vector $v_i = h_\mathrm{CLIP}(I_i)$ by using CLIP image encoder $h_\mathrm{CLIP}$~\cite{radford2021clip}.

\subsection{AD Detection Module}
\label{sec:alg-3}
% Unlike other video events that have distinctive content to detect, such as action and anomalies~\cite{medioni2001event,lai2014video}, the boundaries of AD events are naturally ambiguous. In addition to the gradient passed from the generation loss in the unified model, we leverage the generated AD scripts in our AD event detector to enhance the reliability of detection. To achieve this, we propose two different detection schemes: visual-only and language-guided detection. Fig.~\ref{fig:detector} provides an overview of the proposed detector.

% \vspace*{0.1cm}
% \noindent\textbf{Visual only detection (VOD):} 
We first propose the visual-only detection (VOD) scheme in the upper part of Figure~\ref{fig:overview}, which solely utilizes the enhanced visual features $V$ for AD event detection. It's worth noting that $V$ is also employed for AD script generation. We use a detection-specific encoder $h_\mathrm{det}$, comprising S4 layers~\cite{gu2021s4}, to efficiently derive feature representations for AD event detection. This can be expressed as:
\begin{equation} \label{eq:det_temp}
    X = [x_1, x_2, \ldots, x_N] = h_\mathrm{det}(v_1, v_2, \ldots, v_N),
\end{equation}
where $X \in \mathbb{R}^{N \times D}$ is the sequence of the detection-specific features $x_i \in \mathbb{R}^{D}$ for $i \in \{1, 2, \ldots, N\}$. 

\noindent\textbf{Anchor-based detection:} To facilitate the continuous prediction of AD frames with soft probability assignment, we employ the anchor-based detection framework. This framework casts the problem as a classification task over all potential AD event locations in $\cal V$.
Specifically, during training, we assume there is only one AD event in the clip $\cal V$ with $N$ frames. Then, the number of possible locations of an AD event with length $l$ is $N-l+1$, and the total number of all possible AD events is $M = \sum_{l=1,\ldots,N}{(N-l+1)} = \frac{N(N+1)}{2}$. We define the AD event candidate $c_m$ for $m \in \{1, 2,\ldots, M\}$ as:
\begin{equation}
    C = [c_1, c_2, \ldots, c_{M}]^\top = \begin{bmatrix} 1 & 0 & 0 & \cdots & 0 \\ 0 & 1 & 0 &\cdots &0 \\ \vdots & \vdots & \vdots & \ddots & \vdots \\ 1 & 1 & 1 & \cdots & 1 \end{bmatrix} \in \mathbb{Z}^{M \times N},
\end{equation}
where $1$ indicates that the corresponding frame belongs to an AD event and $0$ represents the opposite.

Then, we obtain the candidate-wise features by aggregating the features at AD event locations of each candidate via candidate pooling, which is defined as:
\begin{equation} \label{eq:candidate_pooling}
    X_c = [x^c_1,  x^c_2, \ldots, x^c_M] = \bar{C}X \in \mathbb{R}^{M \times D},
\end{equation}
Here, $\bar{C} = [\bar{c}_1, \bar{c}_2, \ldots, \bar{c}_M]^\top$ is the normalized candidate from $C$ where $\bar{c}_m = \frac{c_m}{c^\top_m c_M}$ . Note that $c_M$ is the vector of $1$'s. The probability that each candidate matches the ground-truth (GT) AD event $[y_1, y_2, \ldots, y_N]$ is obtained by
\begin{equation} \label{eq:mlp}
    \hat{P}_\mathrm{vod} = [\hat{p}^\mathrm{vod}_1, \hat{p}^\mathrm{vod}_2, \ldots, \hat{p}^\mathrm{vod}_M] = h_\mathrm{MLP}(X_c),
\end{equation}
where $h_\mathrm{MLP}$ is a multi-layer perceptron (MLP). For event candidate $c_m$, $\hat{p}^\mathrm{vod}_m$ informs the probability that it is the AD event.

% Hence, we first obtain GPT2 word embeddings of the AD script and use the temporal encoder for language processing to obtain the refined language features from the word embeddings. Then, the visual features are updated via cross attention with the language features. Then, as in visual detection scheme, we obtain the candidate-wise features via~\eqref{eq:candidate_pooling}, and predict the probability via prediction head.

% \vspace*{0.1cm}
% \noindent\textbf{Ordering relationship estimation:} Most anchor-based object detection techniques perform the bounding box regression to obtain more precise detection results. Similarly, we can perform the event boundary regression. However, different from continuous box boundaries in ordinary object detection, AD event has discrete boundaries. Therefore, instead of regression, we estimate the better matched AD event candidate between two candidates. Specifically, we concatenate the candidate-wise features of two event candidates and predict the better matched one via order prediction head, which is a MLP.

\subsection{Feature Suppression and Generation Module}
\label{sec:alg-4}
In most cases, only a portion of frames in a movie clip $\cal V$ belong to an AD event, while others do not. For reliable AD script generation, we suppress the information of frames that are not part of the AD event, as they may provide unnecessary information to the script generator.

\noindent\textbf{Feature Suppression:} 
To achieve this, we employ the feature suppression module to obtain discriminative visual features based on the AD detection results. Let $\theta \subset \{1, 2, \ldots, M\}$ be the set of candidate indices $i$, where $\hat{p}^\mathrm{vod}_i$ represents the top $k$ probabilities in $\hat{P}_\mathrm{vod}$. Note that $|\theta| = k$. Then, from $V$, we obtain the suppressed features $V_\mathrm{sup}$ by:
% \begin{equation} \label{eq:suppression}
%     V_\mathrm{sup} = 
%     \frac{(\hat{P}^\theta_\mathrm{vod})^\top C^\theta}{C^\theta c_m}V,
% \end{equation}
\begin{equation} \label{eq:suppression}
    V_\mathrm{sup} = \frac{1}{|\theta|}\sum_{i \in \theta}
    \hat{p}^\mathrm{vod}_ic_i\mathbf{1}^\top \odot V,
\end{equation}
where $\mathbf{1}$ denotes a $D$-dimensional vector of ones and $\odot$ is the Hadamard product operator. Hence, it filters out the features corresponding to the frames which are estimated as the non AD event.

% contains the frames belonging to AD event and those not belonging to 
% To generate the reliable AD script, we expect to pass AD-related frames to the AD script generator while suppressing the information of non-AD frames. 

% of $\mathrm{CA^3D}$ aims to generate the reliable script for AD event within a movie clip $\cal V$. Different from the ordinary video captioning task, of which objective is just to describe the visual elements in the scene, it is important to consider the context of the movie in AD scripting, since AD is to describe the story; for example, to generate the AD script of `Dooku fires \textit{again}' in Star Wars, we should know what Dooku did in the previous scenes.

% Due to aforementioned reason, the proposed AD script generator exploits the visual context, which is the movie clip ${\cal V}'=\{I'_1,\ldots,I'_{N'}\}$ with $N'$ frames right before the movie clip $\cal V$, together with $\cal V$. Similar to~\eqref{eq:det_temp}, we obtain $Z' = [z'_1, z'_2, \ldots, z'_{K'}] = f_{\cal V'}(v'_1, v'_2, \ldots, v'_{N'})$ from frame-wise feature vectors $v'_i$ of $I'_i$ extracted by using CLIP encoder $h_\mathrm{CLIP}$. 

% Here, $g_{\cal V'}$ prompt to convey the context for GPT2. 

\noindent\textbf{Script Generation:} For script generation, we follow the basic design of AutoAD~\cite{han2023autoad}. However, we employ visual context information for more reliable AD script generation. Similar to~\eqref{eq:det_temp}, we first obtain the generation-specific context feature $Z' = f_{V'}(V') \in \mathbb{R}^{N' \times D}$ from the enhanced visual context $V'$. Also, we map the suppressed visual features $V_\mathrm{sup}$ to generation-specific features $Z = f_{V}(V_\mathrm{sup}) \in \mathbb{R}^{N \times D}$. We note that $f_{V'}$ and $f_V$ are generation-specific encoders for visual context and visual features, respectively. Both encoders consist of S4 layers. Then, from $Z'$ and $Z$, the frozen GPT2 $f_\mathrm{GPT2}$ generates the probability for each of $K$ tokens in $\cal T$ as
\begin{equation}
\hat{P}_\mathrm{tokens} = [\hat{p}_1, \hat{p}_2, \ldots, \hat{p}_K] = f_\mathrm{GPT2}(Z', Z).
\end{equation}
Note that we use the word embeddings $W$ as additional input to $f_\mathrm{GPT2}$ during training.

%  % \ell_\textrm{gen}(\hat{p}_\mathrm{token}, p_\mathrm{token})
% $\ell_\textrm{focal}$ is soft focal loss which uses IoU score between each candidate and its GT AD location as the ground-truth label. $\ell_\textrm{order}$ is the cross-entropy loss on order prediction. $\ell_\textrm{gen}$ is the cross-entropy loss on AD script prediction. Also, we use the movie clips with no AD event for training the detector so that the proposed algorithm can handle the whole movie. Note that the proposed algorithm is an unified model of detector and generator and it is trained in end-to-end manner. The gradients from $\ell_\textrm{gen}$ affect the parameters in the detector as well, encouraging the detector to find precise AD event location for better quality script generation. 

\subsection{Self-Refinement Module}
\label{sec:alg-6}
Unlike other video events that have distinctive content to detect, such as action and anomalies~\cite{medioni2001event,lai2014video}, the boundaries of AD events are naturally ambiguous. Meanwhile, we believe the detection and generation tasks are complementary to each other in the AD automation system. Thus, in addition to the temporally enhanced visual input, we also propose an optional self-refinement module to leverage the generated AD scripts to further improve the reliability of both detection and generation. 

To this end, we introduce the language-guided detection (LGD) scheme that leverages the AD script $\cal T$ along with the visual information in $V$, providing a complementary cue for AD event detection. In alignment with our AD generation module, we convert the AD script $\cal T$ into a sequence of GPT-2~\cite{radford2019gpt2} word embeddings, $W = [w_1, w_2, \ldots, w_K]$. Similar to~\eqref{eq:det_temp}, we obtain the language features for detection $X_w = [x^w_1, \ldots, x^w_K] = h^{w}_\mathrm{det}(w_1, \ldots, w_K)$ by using another detection-specific encoder $h^{w}_\mathrm{det}$, which has the same architecture as $h_\mathrm{det}$. Then, we use the cross-attention module $h_\mathrm{ca}$ to capture the correlation between language features (serving as keys and values) and visual features (serving as queries) by:
\begin{equation}\label{eq:cross-attn}
X_\mathrm{ca} = h_\mathrm{ca}(X, X_w).
\end{equation}
Here, $X_\mathrm{ca}$ denotes the cross-attended features. Similarly in visual only detection scheme, we predict the detection probability by using $X_\mathrm{ca}$ as input to~\eqref{eq:candidate_pooling} and~\eqref{eq:mlp}. Note that, in language guided scheme, we use the GT AD script $\cal T$ and the generated one $\hat{\cal T}$ for training and inference, respectively.

\noindent\textbf{Iterative Refinement:} Given an entire movie $\cal M$, the movie clips $\{{\cal V}_1, {\cal V}_2, \ldots, {\cal V}_T\}$ and the corresponding visual contexts $\{{\cal V}'_1, {\cal V}'_2, \ldots, {\cal V}'_T\}$ are sampled using a sliding window approach, progressing from the beginning to the end of the movie with a specified step size of $S$. 
For each ${\cal V}_i$ and ${\cal V}'_i$, $\mathrm{CA^3D}$ iteratively refines the detection and generation results. In the initial iteration, $\mathrm{CA^3D}$ utilizes the visual detection scheme to predict the location of AD events. Subsequently, it generates the AD script for ${\cal V}_i$ based on the obtained detection results. Starting from the second iteration, we employ the language guided detection scheme using the AD script generated in the previous iteration to achieve more accurate AD event localization. Additionally, we generate the AD script based on the refined AD event location. This refinement process is iteratively repeated for a predefined number of iterations. Algorithm~\ref{alg:iterative} provides a detailed description of the evaluation process. 

\begin{algorithm}[t]
\caption{$\mathrm{CA^3D}$}
    {\bf Input:} Whole movie $\cal M$
    \begin{algorithmic}[1]
        \State Sample the movie clips $\{{\cal V}_1, {\cal V}_2, \ldots, {\cal V}_T\}$ and the visual contexts $\{{\cal V}'_1, {\cal V}'_2, \ldots, {\cal V}'_T\}$ from $\cal M$;
        \ForAll {$i \in \{1, 2, \ldots, T\}$}
            \State Obtain $V$ and $V'$ from ${\cal V}_i$ and ${\cal V}'_i$ via~\eqref{eq:enhancement}; 
            \ForAll {$j \in \{1, \ldots, J\}$}
                \If {$j = 1$}
                \State
                \parbox[t]{0.79\linewidth}{
                Obtain $X_c$ from $V$ via~\eqref{eq:det_temp} and ~\eqref{eq:candidate_pooling};}
                \State 
                \parbox[t]{0.79\linewidth}{
                Obtain $\hat{P}^{ij}_\mathrm{vod}$ from $X_c$ via~\eqref{eq:mlp}; \Comment{\textit{VOD}}}
                \State
                \parbox[t]{0.79\linewidth}{
                Obtain $V_\mathrm{sup}$ based on $\hat{P}^{ij}_\mathrm{vod}$ via~\eqref{eq:suppression};}
                \State 
                \parbox[t]{0.79\linewidth}{Generate $\hat{\cal T}^j_i$ from $Z$ and $Z'$;  \Comment{\textit{Generation}}}
                \State ${\cal D} \leftarrow \hat{P}^{ij}_\mathrm{vod}$; \quad ${\cal G} \leftarrow \hat{\cal T}^j_i$;
                \ElsIf{$ j > 1 $} \Comment{\textit{Iterative refinement}}
                    \State 
                    \parbox[t]{0.79\linewidth}{Obtain $X_w$ from $\hat{\cal T}^{j-1}_i$;}
                    \State
                    \parbox[t]{0.79\linewidth}{Obtain $X_c$ from $X$ and $X_w$ via~\eqref{eq:cross-attn} and~\eqref{eq:candidate_pooling};}
                    \State
                    \parbox[t]{0.79\linewidth}{Obtain $\hat{P}^{ij}_\mathrm{lgd}$ from $X_c$ via~\eqref{eq:mlp};
                    \Comment{\textit{LGD}}}
                    \State
                    \parbox[t]{0.79\linewidth}{
                    Obtain $V_\mathrm{sup}$ based on $\hat{P}^{ij}_\mathrm{lgd}$ via~\eqref{eq:suppression};}
                    \State 
                    \parbox[t]{0.79\linewidth}{Generate $\hat{\cal T}^j_i$ from $Z$ and $Z'$;  \Comment{\textit{Generation}}}
                    \State ${\cal D} \leftarrow \hat{P}^{ij}_\mathrm{lgd}$; \quad ${\cal G} \leftarrow {\cal T}^j_i$;
                \EndIf
            \EndFor
        \EndFor
    \end{algorithmic}
    {\bf Output:} Detection results $\cal D$, Generation results $\cal G$ 
    \label{alg:iterative}
\end{algorithm}

\subsection{Objective Function}
\label{sec:alg-5}
We define the training loss on the detector outputs as
\begin{equation}
    \ell_\textrm{det} = \ell_\textrm{focal}(\hat{P}_\mathrm{vod}, P_\mathrm{vod}) + \ell_\textrm{focal}(\hat{P}_\mathrm{lgd}, P_\mathrm{lgd})
\end{equation}
where $\hat{P}_\mathrm{vod}$ denotes the detection probabilities of visual only detection and $\hat{P}_\mathrm{lgd}$ denotes those of language guided detection. Also, $P_\mathrm{vod}$ and $P_\mathrm{lgd}$ are their GT probabilities and $\ell_\textrm{focal}$ is the focal loss~\cite{lin2017focal} over binary classes. 

The training loss on the generator outputs is defined as
\begin{equation}
    \ell_\textrm{gen} = \ell_\textrm{ce}(\hat{P}_\mathrm{tokens}, P_\mathrm{tokens}).
\end{equation}
where $\ell_\textrm{ce}$ is the cross-entropy loss. Therefore, the total training loss is defined as $\ell_\textrm{total} = \ell_\textrm{det} + \ell_\textrm{gen}$. It's noteworthy that the parameters of the detector are optimized by $\ell_\textrm{gen}$ as well, owing to the differentiable feature suppression in~\eqref{eq:suppression}. In other words, the detector is incentivized to identify the precise AD event location for improved script generation.

% Then, we conduct the prediction on each movie clip. We define the overlap ratio $r = (N-S)/N$ where $S \leq N$ is the step size of sliding window. Hence, at some frames which are overlapped between adjacent movie clips, we have multiple AD detection results from different movie clips. In such a case, we simply pick the detection results with higher probability as our prediction. 

\section{Experiments}
\label{sec:exp}
% In addition to the results in this section, it is recommended to watch the accompanying video clips in the supplementary materials to see our AD detection and generation results more clearly.

\subsection{Experimental Setup}
\noindent\textbf{Dataset:}
We assess the performance of our model on widely used AD benchmarks. Specifically, \textbf{MADv2~\cite{han2023autoad}} consists of 498 movies, with 488 designated for training and 10 for evaluation. We note that the evaluation split of the MADv2 dataset is identical to the MAD-eval dataset~\cite{han2023autoad}. The dataset comprises pre-extracted frame-wise CLIP features~\cite{radford2021learning} and AD scripts with corresponding timestamps. Additionally, an anonymized version is available, where character names are substituted with the placeholder `someone'. By default, we employ the named version of the MADv2 dataset for both training and evaluation purposes. Moreover, \textbf{AudioVault~\cite{han2023autoad}} encompasses 3.3 million AD events derived from scripts and timestamps across 7,000 movies. Consistent with the methodology outlined in~\cite{han2023autoad}, we exclusively utilize this dataset for pretraining GPT2.

\noindent\textbf{Implementation Details:}
% =================Old text
% For training, the AdamW optimizer is used with a batch size of 128 and a weight decay of 0, and the initial learning rate is $0.0001$. We train the networks for 10 epochs. Also, we perform the scheduled learning according to the cosine-annealing with a linear warm-up of 2,000 steps. We do all experiments using PyTorch and four Tesla V100 GPUs. For text generation, greedy search and beam search are commonly used sampling methods. We stop the text generation when a full stop mark is predicted, otherwise we limit the sequence length to 67 tokens. We use beam search with a beam size of 5 and mainly report results by the top-1 beam-searched outputs, since beam search performs slightly better than greedy search. 
During training, we employ the AdamW optimizer~\cite{loshchilov2017decoupled} with a batch size of 128 and a weight decay of 0. The cosine learning rate scheduler~\cite{gotmare2018closer} is initialized at $0.0001$, and the networks are trained for 10 epochs with a linear warm-up period of 2,000 steps (equivalent to 0.75 epoch). The experiments are conducted using PyTorch~\cite{imambi2021pytorch} on four Tesla V100 GPUs. For text generation, we utilize beam search~\cite{kumar2013beam} with a beam size of 5, reporting results based on the top-1 outputs from the beam search by default. Text generation stops upon predicting a full stop mark; otherwise, we limit the sequence length to 67 tokens, as introduced in~\cite{han2023autoad}. Additionally, we set $N=32$, $N'=64$, and $K=36$. Further details can be found in the supplementary materials.

\begin{table}[t]
    \caption
    {
    Comparison of AD detection results on MADv2 dataset. Here, `V' and `AD' denote the visual inputs and previous AD context inputs, respectively.
    }
    \vspace*{-0.1cm}
    \centering
    \resizebox{0.9\linewidth}{!}{
    \footnotesize
    \begin{tabular}{@{} l c c c c c @{} }
        \toprule
        Algorithm                  &  Inputs   & Precision & Recall &  F1    \\
        \midrule
        Random                     &   V       &  29.3    &  43.4  &  35.1  \\
        TriDet~\cite{shi2023tridet}&   V       &  52.1    &  76.5  &  61.9  \\  
        MIGCN~\cite{zhang2021migcn}&   V + AD  &  69.8    &  69.8  &  69.8  \\
        LGI~\cite{mun2020local}    &   V + AD  &  \textbf{71.4}    &  71.4  &  \textbf{71.4}  \\
        $\mathrm{CA^3D}$           &   V       &  55.2    &  \textbf{80.1}  &  65.3  \\
        \bottomrule
    \end{tabular}
   \label{table:detection}
   }
   %\vspace*{-0.1cm}
\end{table}

% ===========Original text
% \noindent\textbf{MADv2~\cite{}}: It is a dataset for audio description. It contains 498 movies, which are split into 488 training and 10 evaluation movies. For each frame in the movies, feature vector, which is extracted via CLIP~\cite{} visual encoder, is provided. Also, for each AD event, AD script and its time stamps are provided.

% \vspace*{0.2cm}
% \noindent\textbf{AudioVault~\cite{}}: It contains 3.3M AD events with the script and tims stamps from 7,000 movies which are not included in the MADv2 dataset. As in~\cite{}, we use this dataset for text-only pretraining of GPT2 in our AD generator.

% \noindent\textbf{LSMDC2019~\cite{}}: It is a dataset for multi-sentence description. It contains 128,085 movie clips from 200 movies and the descriptions for each movie clip.   

\noindent\textbf{Evaluation Metrics:} As the proposed $\mathrm{CA^3D}$ is the first unified system which tackles both AD detection and generation tasks, we evaluate two tasks separately to demonstrate the effectiveness. To assess the performance of AD event detection, we compute \textit{Precision}, \textit{Recall}, and \textit{F1} scores. An identified AD event throughout the entire movie is deemed correct if its intersection-over-union (IoU) ratio with the ground truth surpasses $0.1$. For the evaluation of AD script generation, we employ Rouge-L~\cite{lin2004rouge} and CIDEr scores~\cite{vedantam2015cider} metrics, consistent with previous works~\cite{han2023autoad,han2023autoad2}.  For each GT AD script, we compute both metrics with the generated AD script whose detected temporal location is closest to it. 
% The Rouge-L and CIDEr scores are computed using the pycocoevalcap library~\cite{lin2014microsoft}.
% Rouge-L is a recall-based metric calculated with the length of the longest common subsequence. CIDEr captures consensus by applying Term Frequency-Inverse Document Frequency (TF-IDF) weighting for each n-gram.
% \vspace*{0.2cm}
% \noindent\textbf{AD script generation:} 

% \noindent\textbf{WebVid~\cite{}}: It is a large video-caption dataset. It contains 2.5M short videos with text descriptions, which are sampled from stock footage websites. We use this dataset for pretraining the temporal encoder in the script generator, as in~\cite{}.

\subsection{Main Results}
\noindent\textbf{AD Event Detection:}
Table~\ref{table:detection} presents the performance of AD event detection on the MADv2 dataset. For comparison, we include scores of random estimation as the lower bounds while the performance of LGI~\cite{mun2020local} and MIGCN~\cite{zhang2021migcn} as the upper bounds performance. Specifically, LGI~\cite{mun2020local} and MIGCN are the state-of-the-art methods for the Natural Language Video Grounding (NLVG) task~\cite{gao2017tall, anne2017localizing, wang2022negative, sun2022you}, which aims to accurately locate the video moment semantically corresponding to a specific linguistic query. In Table~\ref{table:detection}, they are evaluated using ground-truth AD locations and scripts. As LGI and MIGCN localizes one AD event per linguistic query, the values for \textit{False Positive} and \textit{False Negative} are identical. So the value of \textit{Precision}, \textit{Recall} and \textit{F1} for LGI~\cite{mun2020local} are the same. Besides, we also include TriDet~\cite{shi2023tridet} as one baseline which is the state-of-the-art method for action detection.

\begin{table}[t]
    \caption
    {
    Comparison of AD generation results on the named version of MADv2 dataset. 
    }
    \vspace*{-0.1cm}
    \centering
    \resizebox{\linewidth}{!}{
    \footnotesize
    \begin{tabular}{@{} l c c c c c @{} }
        \toprule
        Algorithm                               & Pretrain & Inputs & AD location & Rouge-L & CIDEr \\
        \midrule
        SwinBERT~\cite{lin2022swinbert}                       & AudioVault & V      &  \checkmark & 8.5     &  6.7  \\
        AutoAD~\cite{han2023autoad}             & WebVid + AudioVault & V      & \checkmark  & 9.9     & 10.0  \\
        AutoAD~\RomNum{2}~\cite{han2023autoad2} & WebVid + AudioVault & V      & \checkmark  & 9.7     & 10.0  \\
        $\mathrm{CA^3D}$          & AudioVault & V      & \checkmark  & \textbf{11.3}    & \textbf{10.8}  \\
        $\mathrm{CA^3D}$          & AudioVault & V      &             & \textbf{11.3}    &  9.4  \\
        \midrule
        AutoAD~\cite{han2023autoad}            & WebVid + AudioVault & V + AD & \checkmark  & 13.9    & 19.0  \\
        AutoAD~\RomNum{2}~\cite{han2023autoad2} & WebVid + AudioVault & V + Char.  & \checkmark  & 13.1    & 19.2  \\
        $\mathrm{CA^3D}$          & AudioVault & V + AD & \checkmark  & \textbf{14.0}    & \textbf{20.4}  \\
        $\mathrm{CA^3D}$          & AudioVault & V + AD &             & 13.5    & 17.7  \\
        
        \bottomrule
        \end{tabular}
   \label{table:generation_named}
   }
   %\vspace*{-0.1cm}
\end{table}

\begin{table}[t]
    \caption
    {
    Comparison of AD generation results on the unnamed version of MADv2 dataset. 
    }
    \vspace*{-0.1cm}
    \centering
    \resizebox{\linewidth}{!}{
    \footnotesize
   \begin{tabular}{@{} l c c c c @{} }
        \toprule
        Algorithm         & Inputs  & AD location & Rouge-L & CIDEr \\
        \midrule
        AutoAD~\cite{han2023autoad}           & V + AD  & \checkmark    & 15.9   & 14.5  \\
        $\mathrm{CA^3D}$         & V + AD  & \checkmark    & \textbf{16.2}   & \textbf{14.8}  \\
        $\mathrm{CA^3D}$         & V + AD  &               & 15.8   & 13.9  \\
        \bottomrule
        \end{tabular}

   \label{table:generation_unnamed}
   }
   %\vspace*{-0.1cm}
\end{table}

In Table~\ref{table:detection}, the proposed $\mathrm{CA^3D}$ demonstrates superior performances compared to TriDet~\cite{shi2023tridet} across all metrics. Furthermore, it achieves competitive results with LGI~\cite{mun2020local} and MIGCN~\cite{zhang2021migcn}, even though LGI~\cite{mun2020local} benefits from access to subsequent and preceding AD ground-truth scripts and visual information. It's noteworthy that LGI~\cite{mun2020local} achieves an F1 score of $71.4$, underscoring the challenging nature of the AD detection task even with access to ground truth AD scripts. Importantly, $\mathrm{CA^3D}$ outperforms LGI~\cite{mun2020local} in \textit{Recall} by $8.7\%$. While the \textit{Precision} may be lower than the upper bound, a high \textit{Recall} is advantageous in the context of AD automation. This is because \textit{False Positive} detection can be addressed during post-processing, while \textit{False Negatives} are challenging to rediscover in long movies.

\noindent\textbf{AD Script Generation:} Table~\ref{table:generation_named} presents a comparison of the performance of AD script generation on the named version of the MADv2 dataset~\cite{han2023autoad2}. We include AutoAD~\cite{han2023autoad}, AutoAD~\RomNum{2}~\cite{han2023autoad2} and SwinBERT~\cite{lin2022swinbert} as baselines which are the state-of-the-art methods in AD generation and video captioning. Please note they generate AD scripts by using the ground truth AD locations. When all methods are provided with the oracle AD location, the proposed algorithm exhibits consistent better performance for both visual-only input and visual plus linguistic context (either AD context or character name). Remarkably, $\mathrm{CA^3D}$ achieves a CIDEr score $1.2$ higher than AutoAD~\RomNum{2}, which leverages additional metadata and employs an additional network for character name recognition. Moreover, $\mathrm{CA^3D}$ still maintains competitive performances with AutoAD and AutoAD~\RomNum{2} when generating AD scripts within the detected AD event windows (without GT AD location). Additionally, Table~\ref{table:generation_unnamed} provides results on the unnamed version of the MADv2 dataset, where $\mathrm{CA^3D}$ achieves the best scores across all metrics. The lower performances of TriDet and SwinBERT suggest that the simple adoption of SOTA methods in other applications is not practically useful in solving AD automation. Notably, unlike AutoAD and AutoAD~\RomNum{2}, $\mathrm{CA^3D}$ is \textbf{not} pretrained on the \textbf{WebVid} dataset~\cite{bain2021frozen}, which contains \textbf{2.5M} short video-text pairs. The proposed algorithm not only establishes new state-of-the-art performance but also demonstrates significant data efficiency.

\begin{table}[t]
    \caption
    {
    Ablation studies for modules in the detector on the MADv2 dataset. 
    }
    \vspace*{-0.1cm}
    \centering
    \resizebox{0.95\linewidth}{!}{
    \footnotesize
    \begin{tabular}{@{} l c c c c c @{} }
        \toprule
        Method     & Enhancement  & Anchor        & Precision & Recall & F1 \\
        \midrule
        \RomNum{1} &  \checkmark  &               & 51.9 & 77.2  & 62.1  \\
        \RomNum{2} &              & \checkmark    & 53.3 & 79.4  & 64.5 \\
        \RomNum{3} &  \checkmark  & \checkmark    & 54.4 & 82.2  & 65.4 \\
        \bottomrule
    \end{tabular}
   \label{table:ablation_detector}
   }
   %\vspace*{-0.1cm}
\end{table}

\begin{table}[t]
    \caption
    {
    Ablation studies for modules in the generator on the MADv2 dataset. 
    }
    \vspace*{-0.1cm}
    \centering
    \resizebox{0.95\linewidth}{!}{
    \footnotesize
    \begin{tabular}{@{} l c c c c @{} }
        \toprule
        Method     &  Enhancement   & Suppression   & Rouge-L & CIDEr \\
        \midrule
        \RomNum{1} &  \checkmark    &               & 10.8    & 8.9   \\
        \RomNum{2} &                & \checkmark    & 11.0    & 9.2   \\
        \RomNum{3} &  \checkmark    & \checkmark    & 11.3    & 9.4   \\
        \bottomrule
    \end{tabular}
   \label{table:ablation_generator}
   }
   %\vspace*{-0.1cm}
\end{table}

\begin{figure}[t] 
    \begin{subfigure}[t]{.49\linewidth}
    \centering
    \includegraphics[width=0.99\textwidth]{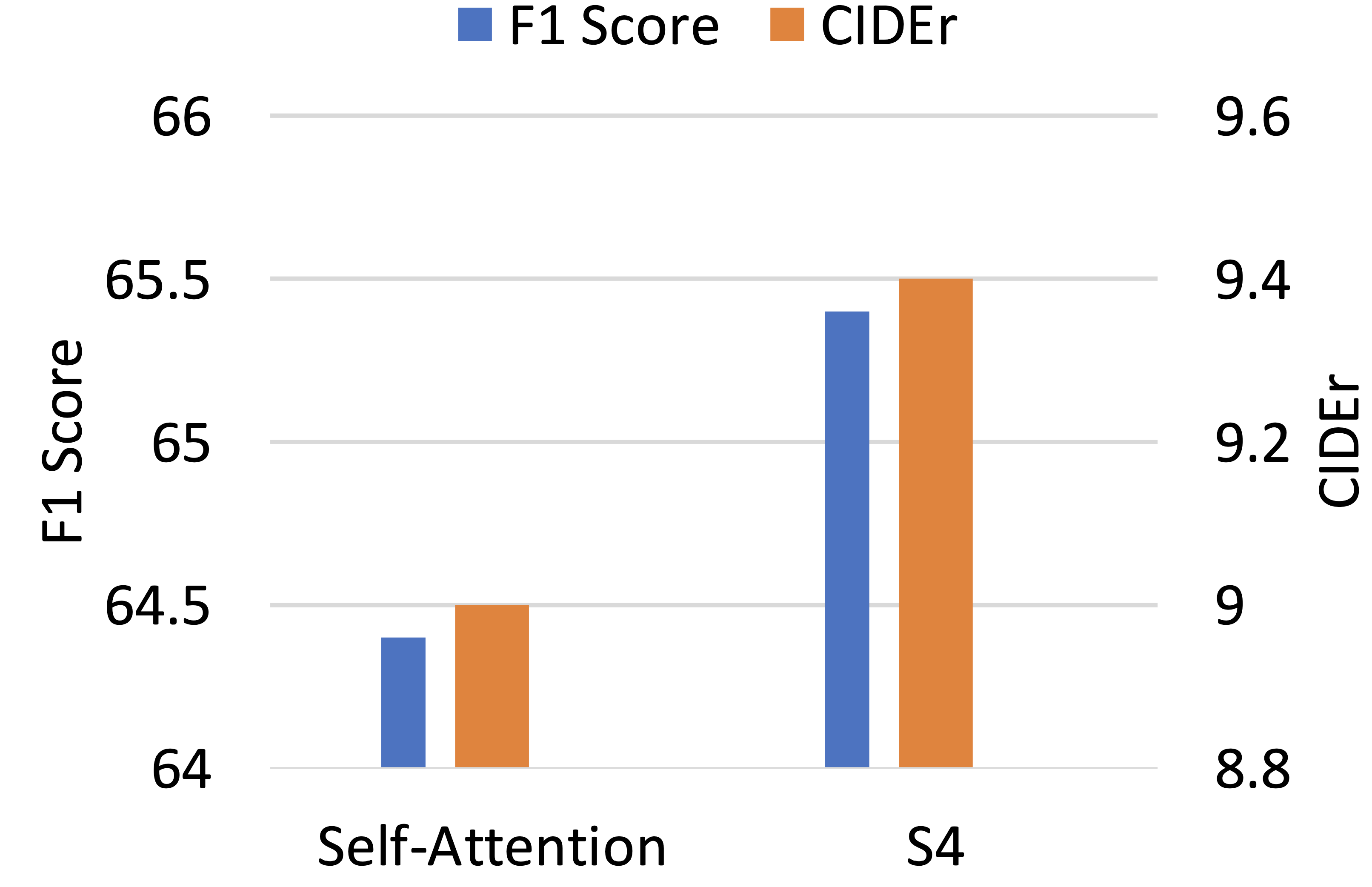}
    \caption{}
    \label{fig:attn_s4}
    \end{subfigure}
    \begin{subfigure}[t]{.49\linewidth}
    \centering
    \includegraphics[width=0.99\textwidth]{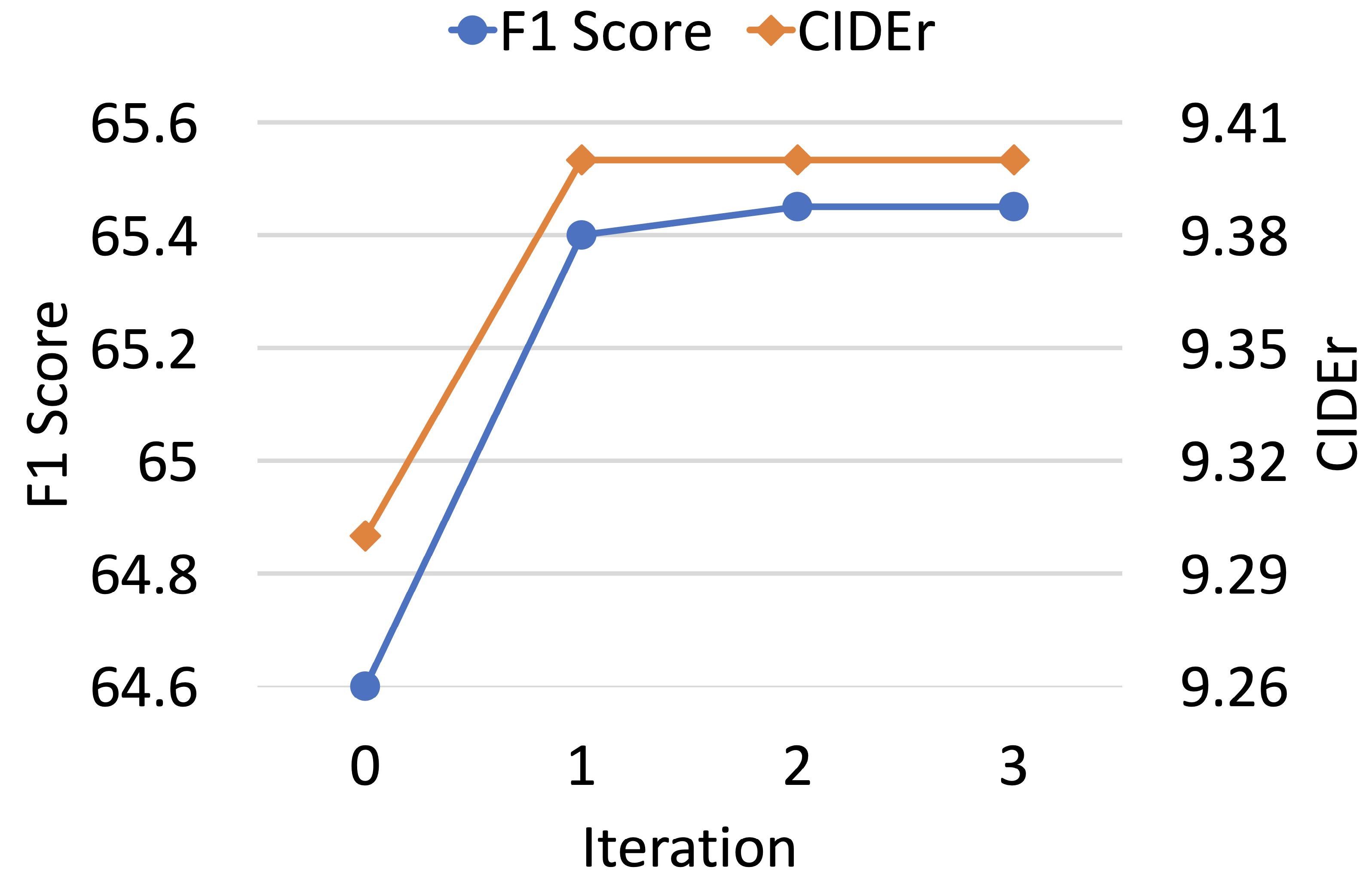}
    \caption{}
    \label{fig:iterative_detection}
    \end{subfigure}
    \caption{(a) Comparison of AD generation and detection performances at different iterations, (b) Comparison of AD generation and detection performances at different iterations.}
\end{figure}

\subsection{Ablation Study}
% Due to space constraints, additional ablation studies and in-depth analyses are provided in the supplementary materials.
% , offering a more comprehensive exploration of the proposed model and its components.

\noindent\textbf{Context-aware Temporal Feature Enhancement:} In Tables~\ref{table:ablation_detector} and~\ref{table:ablation_generator}, we assess the effectiveness of the proposed feature enhancement module. In this work, we advocate the use of a simple S4 module to integrate longer temporal cues into visual features with concise content, which is more efficient than previous architecture. From Tables~\ref{table:ablation_detector} and~\ref{table:ablation_generator}, it is evident that the feature enhancement module enhances the performance of $\mathrm{CA^3D}$ in both detection and generation tasks. Figure~\ref{fig:attn_s4} shows that the proposed feature enhancement module achieves better performances with S4 layers than standard self-attention layers. We note that the computational complexity of S4 is $\mathcal{O}(n)$ while the one of transformer layer is $\mathcal{O}(n^2)$, where $n$ is the sequence length. The linear complexity of S4 enables the success of our proposed temporal feature enhancement module in the long-form video domain.

\begin{figure*}[t] 
    \centering
    \includegraphics[width=0.9\linewidth]{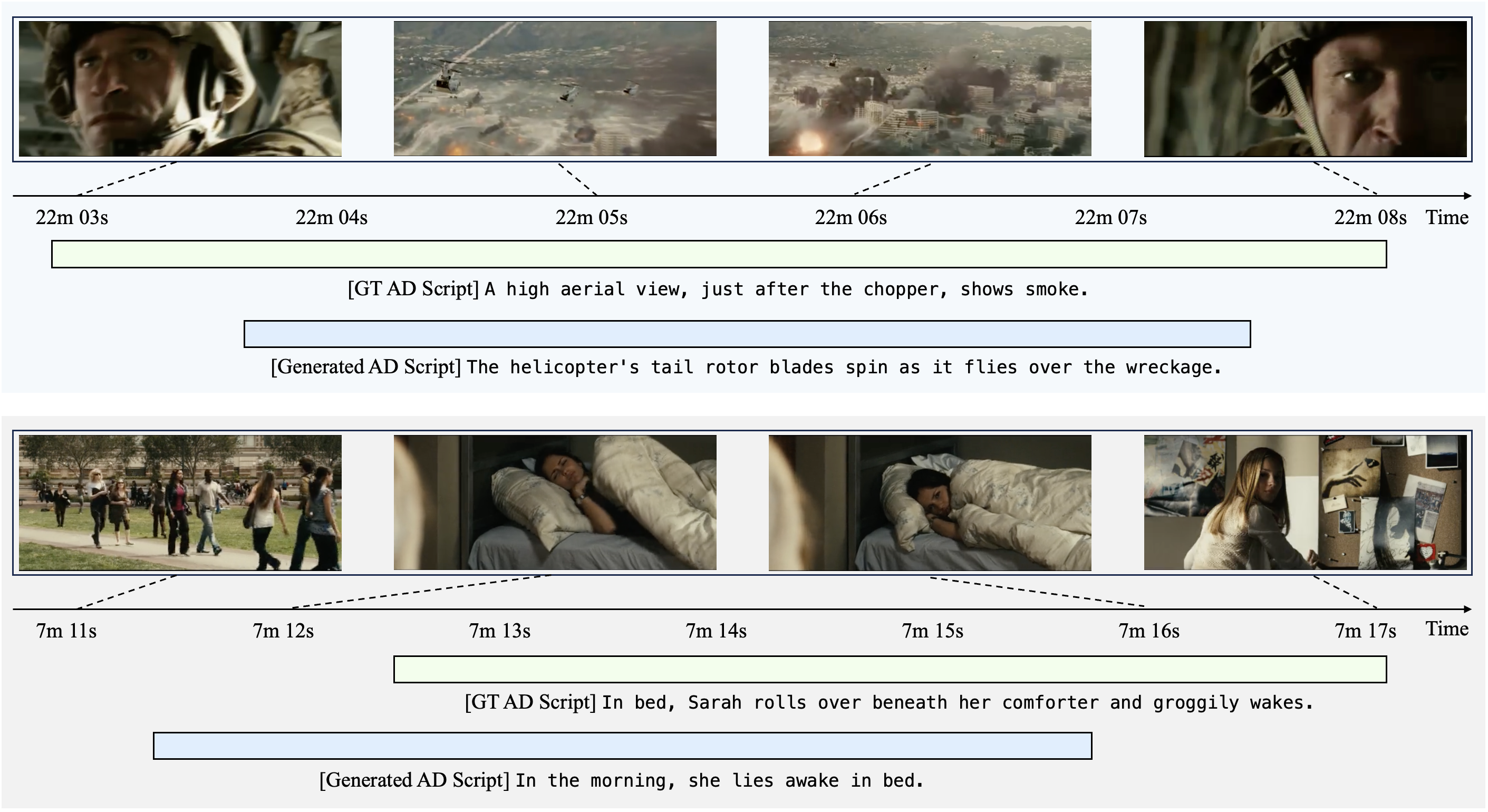}
    \caption{Examples of AD detection and generation results on the MADv2 test set.}
    %\vspace*{-0.1cm}
    \label{fig:examples}
    \vspace*{-0.3cm}
\end{figure*}

\noindent\textbf{Anchor-based Prediction Head:} In Table~\ref{table:ablation_detector}, we conduct an ablation study on the proposed anchor-based prediction head in the AD event detector. Method~\RomNum{1} does not utilize AD event candidates and directly predicts whether each frame belongs to an AD event from frame-wise features. This can be considered a segmentation-based AD detection baseline. By comparing Method~\RomNum{1} and Method~\RomNum{3} in Table~\ref{table:ablation_detector}, it is evident that the proposed anchor-based prediction head significantly improves the detection results, with improvements of $+2.5$, $+5.0$, and $+3.3$ in \textit{Precision}, \textit{Recall}, and \textit{F1}.

\noindent\textbf{Feature Suppression:} In Table~\ref{table:ablation_generator}, we conduct an ablation study on the proposed feature suppression module in the AD script generator. Method~\RomNum{1} does not utilize the feature suppression module, using $V$ instead of $V_{\mathrm{sup}}$ in Algorithm~\ref{alg:iterative} as the baseline model. By comparing Method~\RomNum{1} and Method~\RomNum{3} in Table~\ref{table:ablation_generator}, it is evident that the feature suppression module brings clear benefits, improving both Rouge-L and CIDEr by $0.5$. This result shows that it is crucial to eliminate visual information not corresponding to the AD events.

\noindent\textbf{Self-refinement:} Figure~\ref{fig:iterative_detection} outlines the detection and generation scores on the MADv2 dataset at each iteration. We employ the visual detection scheme for the initial prediction (iteration 0) and then use the language-guided detection in later iterations. The detection performance improves as the iteration goes on, indicating that the generated scripts contribute to enhanced AD event detection. Moreover, the proposed algorithm achieves higher generation scores when opt-in the self-refinement module which enables more accurate detection results. The performance saturates after one iteration, which suggests the effectiveness of the refinement module. These results underscore the complementary nature of AD event detection and AD script generation within the proposed automation system. 

% \begin{figure*}[t]
%     \centering
%     \includegraphics[width=0.9\linewidth]{figure/dummy.png}
%     \caption{Examples of AD detection results on the MADv2 evaluation dataset.}
%     %\vspace*{-0.1cm}
%     \label{fig:examples_generation}
%     %\vspace*{-0.1cm}
% \end{figure*}

% \begin{table}[t]
%     \caption
%     {
%     Comparison of AD generation and detection performances at different iterations. 
%     }
%     \vspace*{-0.1cm}
%     \centering
%     \resizebox{0.5\linewidth}{!}{
%     \footnotesize
%     \begin{tabular}{@{} l c c c c c c @{} }
%         \toprule
%         & \multicolumn{3}{c}{Detection} & \multicolumn{2}{c}{Generation}\\
%         \cmidrule(lr){2-4} \cmidrule(lr){5-6}
%         Iteration & Precision & Recall & F1 & Rouge-L & CIDEr\\
%         \midrule
%         0 &  53.6 & 81.2  & 64.6 & 11.3 & 9.3  \\
%         1 &  54.4 & 82.2  & 65.4 & 11.3 & 9.4  \\
%         \bottomrule
%     \end{tabular}
%    \label{table:iterative_detection}
%    }
%    \vspace*{-0.3cm}
% \end{table}

\subsection{Visualizations}
% Figure~\ref{fig:examples} displays some examples of AD detection and generation results on the MADv2 test set. We see that $\mathrm{CA^3D}$ yields decent results for both detection and generation. We note that AD event location is subjective and somewhat ambiguous, as different individuals may prefer to define it differently (e.g., starting from the beginning or the middle of a scene). Also, in the MADv2 dataset, AD event timestamps are not always correct, as shown in the upper part of Figure~\ref{fig:examples}. Hence, even though the detected AD event locations are not perfectly aligned with the GT, they are precise enough for facilitating AD process. Similarly, the generated AD scripts are not exactly the same with the GT AD scripts. However, they effectively describe the scenes, showing one promising step towards the complete AD automation. 
Figure~\ref{fig:examples} shows the examples of AD detection and generation results on the MADv2 test set. $\mathrm{CA^3D}$ yields satisfactory outcomes for both detection and generation tasks. We note that the determination of AD event locations is subjective and somewhat ambiguous, as different individuals may prefer different locations (e.g., commencing from the beginning or the middle of a scene). Moreover, within the MADv2 dataset, some AD event timestamps are not accurate, as illustrated in the upper part of Figure~\ref{fig:examples}. Even though the detected AD event locations may not align perfectly with the ground truth, $\mathrm{CA^3D}$ provides useful results which are precise enough to facilitate the AD process. Similarly, while the generated AD scripts may not exactly match the GT AD scripts, they describe the scenes properly.

% showcasing the proficiency of our detector in identifying AD event locations. It is important to acknowledge that AD event location is subjective and somewhat ambiguous, as different individuals may prefer to define it differently (e.g., starting from the beginning or the middle of a scene). In the MADv2 dataset, there is no consistent rule for determining the starting point as long as the AD event location is within the scene boundary. Consequently, even if the detected location does not perfectly align with the ground truth (GT) location, it does not adversely affect the user experience as long as it is contained within the scene. Furthermore, in certain instances, the detected AD location is more precise than the GT AD location.
 
% Figure~\ref{fig:examples} also presents examples of our generation results on the MADv2 evaluation set. Despite the generated scripts not being identical to the ground truth (GT) scripts, the proposed algorithm effectively describes the scenes. Additionally, in Figure~\ref{fig:examples_generation}, it is evident that the proposed algorithm produces a higher-quality AD script compared to the GT script.

\vspace{-0.2cm}
\section{Conclusion}
In this paper, we tackle the AD automation through an unified system, $\mathrm{CA^3D}$, which identifies the AD events and simultaneously generates the corresponding AD scripts on cinematic data. $\mathrm{CA^3D}$ involves a novel temporal enhancement module to first expand the temporal horizon of the input, an anchor-based AD detector working seamlessly with the feature suppression module to extract discriminative representation and a self-refinement module to further boost the performance. $\mathrm{CA^3D}$ operates directly on long-form videos, establishing the new state-of-the-art performance in the AD automation.

%%%%%%%%% REFERENCES
{\small
\bibliographystyle{ieee_fullname}
\bibliography{main}
}

\end{document}